\PassOptionsToPackage{colorlinks=true,allcolors=blue}{hyperref}
\documentclass[11pt]{article}

\usepackage{acl}

\usepackage{times}
\usepackage{latexsym}

\usepackage[T1]{fontenc}

\usepackage[utf8]{inputenc}

\usepackage{microtype}

\usepackage{inconsolata}
\usepackage{booktabs}

\usepackage[edges]{forest}
\usepackage{adjustbox}
\usepackage[most]{tcolorbox}

\usepackage[table]{xcolor}
\definecolor{taskcolor}{RGB}{226,229,235}
\definecolor{benchcolor}{RGB}{223,235,229}
\definecolor{methodcolor}{RGB}{255,253,246}
\usepackage{multirow}
\usepackage{harveyballs}
\usepackage{pifont}

\usepackage[capitalize,noabbrev]{cleveref}
\crefname{section}{\S}{\S}

\usepackage{graphicx}
\usepackage{epigraph}
\usepackage{varwidth}
\usepackage{ragged2e}
\usepackage{makecell}

\usepackage{tikz}

\newcommand{\emark}{%
\tikz[baseline=-0.55ex,scale=0.13]{
  \fill[violet!80!black] (0,0) -- (90:1)
    arc[start angle=90,end angle=45,radius=1] -- cycle;
  \draw[violet!80!black,line width=0.45pt] (0,0) circle (1);
}}

\title{Contextual Causality with Large Language Models: A Survey}

\author{
  \textbf{Yiheng Zhao\textsuperscript{1,\textdagger}},
  \textbf{Jun Yan\textsuperscript{1}},
  \textbf{Chengming Hu\textsuperscript{2}}
  \\
  \textsuperscript{1}Concordia University, Montreal, Canada \\
  \textsuperscript{2}Mcgill University, Montreal, Canada  \\
  \small{
  \textsuperscript{\textdagger}\textbf{Correspondence:}
  \href{mailto:yiheng.zhao@mail.concordia.ca}{yiheng.zhao@mail.concordia.ca}
}
}

\definecolor{semanticcolor}{RGB}{68,114,196}
\definecolor{interventioncolor}{RGB}{169,134,167}
\definecolor{countercolor}{RGB}{224,188,201}

\forestset{
  semanticbranch/.style={
    draw=semanticcolor,
    for descendants={draw=semanticcolor},
  },
  interventionbranch/.style={
    draw=interventioncolor,
    for descendants={draw=interventioncolor},
  },
  counterbranch/.style={
    draw=countercolor,
    for descendants={draw=countercolor},
  },
  topedge/.style={
  edge path={
    \noexpand\path[\forestoption{edge}]
    (!u.parent anchor) -| ([xshift=-6pt].child anchor) -- (.child anchor)\forestoption{edge label};
  },
}
}

\begin{document}
\maketitle
\begin{abstract}
Understanding contextual causality is critical for large language models (LLMs), as it enables them to accurately identify causal relations in specific situations and support more reliable decision-making. Despite its significance, a systematic exploration of contextual causality with LLMs is still lacking. To fill this gap, we present a comprehensive survey on this topic. In this survey, we first propose a taxonomy of contextual causality, consisting of semantic, intervention, and counterfactual causality, and characterize each category by its core causal question, required model capabilities, representative tasks, and practical uses in causality analysis. We then analyze existing studies and discuss their key limitations. Finally, we examine the gaps between current benchmarks and real-world needs and outline promising directions for future research. Our goal is to clarify the research landscape of contextual causality with LLMs, emphasize its importance, and highlight promising future directions.
\end{abstract}

\section{Introduction}

\begin{center}
\begin{minipage}{0.73\linewidth}
\itshape
Human knowledge and human power meet in one; for where the cause is not known the effect cannot be produced.\hfill --- Francis Bacon
\end{minipage}
\end{center}

\noindent Causality, an important aspect of intelligence, can be broadly divided into non-contextual and contextual forms \cite{lake2017building, halpern2016actual}. Non-contextual causality refers to general causal relations between variables, such as smoking increasing the risk of lung cancer. In contrast, contextual causality concerns particular causal relations grounded in concrete situations, such as whether a specific patient's lung cancer is caused by long-term smoking. In natural language settings, contextual causality is grounded in a given natural language situation, involving causal relations either among events or between actions and outcomes. 

Understanding such causality is essential for large language models (LLMs), as it enables them to accurately identify causal relations and better estimate the outcomes of actions, thereby supporting more reliable decision-making \cite{drury2022survey, wang2024causalbench, yang2022survey, halpern2016actual, lake2017building}. This is especially important in high-stakes domains \cite{richens2020improving, wu2024causal}. For example, in personalized medical diagnosis, LLMs need to identify the causal relations underlying a patient's symptoms, medical history, and clinical records to identify factors driving the current condition, anticipate the effects of different treatments, and support more reliable decisions. 

\begin{table*}[!t]
\centering
\footnotesize
\setlength{\tabcolsep}{2pt}
\renewcommand{\arraystretch}{1.08}
\begin{tabular}{lccccc}
\toprule
\multirow{2}{*}{\textbf{Survey}} 
& \multicolumn{3}{c}{\textbf{Contextual Causality}} 
& \multicolumn{2}{c}{\textbf{Non-Contextual Causality}} \\
\cmidrule(lr){2-4} \cmidrule(lr){5-6}
& \textbf{Semantic} 
& \textbf{Intervention} 
& \textbf{Counterfactual} 
& \textbf{Causal Discovery} 
& \textbf{Causal Inference} \\
\midrule

\citep{wan2024large} 
& \textcolor{orange!75!black}{\ding{55}}  & \textcolor{orange!75!black}{\ding{55}} & \textcolor{orange!75!black}{\ding{55}}
& \textcolor{teal}{\ding{51}} & \textcolor{orange!75!black}{\ding{55}} \\

\citep{liu2025large} 
& \textcolor{orange!75!black}{\ding{55}} & \textcolor{orange!75!black}{\ding{55}} & \textcolor{orange!75!black}{\ding{55}} 
& \textcolor{teal}{\ding{51}} & \textcolor{teal}{\ding{51}} \\

\citep{ma2025causal} 
& \textcolor{orange!75!black}{\ding{55}} & \textcolor{orange!75!black}{\ding{55}} & \textcolor{orange!75!black}{\ding{55}} 
& \textcolor{orange!75!black}{\ding{55}}  & \textcolor{teal}{\ding{51}} \\

\citep{yu2025causaleval} 
& \textcolor{orange!75!black}{\ding{55}} & \textcolor{orange!75!black}{\ding{55}} & \textcolor{orange!75!black}{\ding{55}} 
& \textcolor{teal}{\ding{51}} & \textcolor{teal}{\ding{51}} \\

\citep{zhou2025emerging} 
& \emark & \textcolor{orange!75!black}{\ding{55}} & \textcolor{orange!75!black}{\ding{55}} 
& \emark & \emark \\

\citep{bazgir2025causal} 
& \textcolor{orange!75!black}{\ding{55}} & \textcolor{orange!75!black}{\ding{55}} & \textcolor{orange!75!black}{\ding{55}} 
& \textcolor{teal}{\ding{51}} & \textcolor{teal}{\ding{51}} \\

\citep{li2025survey} 
& \textcolor{orange!75!black}{\ding{55}} & \textcolor{orange!75!black}{\ding{55}} & \textcolor{orange!75!black}{\ding{55}} 
& \emark & \emark \\

\citep{wang2024survey} 
& \textcolor{orange!75!black}{\ding{55}} & \textcolor{orange!75!black}{\ding{55}} & \textcolor{teal}{\ding{51}}$_{\mathrm{CG}}$
& \textcolor{orange!75!black}{\ding{55}}  & \textcolor{orange!75!black}{\ding{55}}  \\

\citep{cheng2025survey} 
& \emark & \textcolor{orange!75!black}{\ding{55}} & \textcolor{orange!75!black}{\ding{55}}
& \textcolor{orange!75!black}{\ding{55}}  & \textcolor{orange!75!black}{\ding{55}}  \\

\textbf{Ours}
& \textcolor{teal}{\ding{51}} & \textcolor{teal}{\ding{51}} & \textcolor{teal}{\ding{51}}$_{\mathrm{CR}}$
& \textcolor{orange!75!black}{\ding{55}}  & \textcolor{orange!75!black}{\ding{55}}  \\
\bottomrule
\end{tabular}
\caption{Comparison with methodology-oriented causality surveys. \textcolor{teal}{\ding{51}}~indicates systematic coverage, 
\protect\emark~indicates coverage of only a few relevant studies, 
and \textcolor{orange!75!black}{\ding{55}}~indicates no systematic coverage. \citep{wang2024survey} mainly covers counterfactual generation (CG), whereas ours covers broader counterfactual reasoning (CR).}
\label{tab:survey_comparison}
\end{table*}

Despite its importance, a systematic overview of contextual causality with LLMs remains lacking. Although several surveys have recently discussed the role of LLMs in causality, most of them primarily focus on non-contextual causality, including causal discovery and causal inference with LLMs \cite{wan2024large, liu2025large, ma2025causal, yu2025causaleval, zhou2025emerging, bazgir2025causal, li2025survey}. There are also task-specific surveys on problems related to contextual causality, such as counterfactual generation (CG) with LLMs and event causality identification (ECI) \cite{wang2024survey, cheng2025survey}. However, these surveys cover only a limited portion of the broader landscape of contextual causality. In particular, the ECI survey reviews the development of ECI and includes only limited coverage of recent studies on ECI with LLMs. In addition, some surveys discuss broader thematic topics such as commonsense causality, causal uncertainty, and benchmark analysis \cite{cui2024odyssey, cui2025uncertainty, yang2024critical}. Therefore, despite its importance, a systematic overview of contextual causality with LLMs remains lacking. To fill this gap, we conduct a comprehensive survey on this topic.

In this survey, we first introduce a taxonomy of contextual causality, motivated by causal analysis needs, and organize it into semantic, intervention, and counterfactual causality. For each category, the taxonomy characterizes the core causal question, required model capabilities, representative tasks, and practical uses in causality analysis (\cref{sec:taxonomy}). We then analyze existing studies, covering both evaluation-oriented and method-oriented research, and discuss their limitations (\cref{sec:method}). Furthermore, unlike existing surveys that examine whether these benchmarks can truly assess LLMs' contextual causality abilities or merely test their ability to retrieve knowledge \cite{yang2024critical}, we focus on the gap between them and real-world demands (\cref{sec:benchmark}). Lastly, we suggest promising directions, including real-world-oriented benchmark construction, more comprehensive and fine-grained evaluation, post-training data construction, and greater attention to intervention and counterfactual causality (\cref{sec:future}). The comparison with existing surveys is summarized in Table \ref{tab:survey_comparison}. The resulting taxonomy of contextual causality is provided in Table \ref{tab:taxonomy}.

\begin{table*}[!t]
\centering
\footnotesize
\setlength{\tabcolsep}{2pt}
\renewcommand{\arraystretch}{1.15}
\begin{tabular}{llll}
\toprule
Core Causal Question & 
Capability & 
Task & 
Practical Use \\
\midrule

\rowcolor{gray!30}
\multicolumn{4}{c}{\textit{Semantic Causality}} \\

\multirow{6}{*}{\makecell[l]{What causal information\\is expressed or implied\\in context?}}
& \multirow{4}{*}{Identification}
& Event Causality Identification 
& Causal Graph Construction \\

& 
& Causal Judgment 
& Causal Link Refinement \\

& 
& Causal Attribution
& Responsibility Assignment \\

& 
& Causal Detection and Span Extraction
& \makecell[l]{Causal Filtering and\\Evidence Grounding} \\

\cmidrule(l){2-4}

& Explanation
& Causal Explanation Generation
& Causal Mechanism Explanation \\

\cmidrule(l){2-4}

& Strength Estimation
& Causal Strength Assessment
& Quantifying Causal Influence \\

\midrule

\rowcolor{gray!30}
\multicolumn{4}{c}{\textit{Intervention Causality}} \\

\makecell[l]{What outcome would occur\\under a given intervention\\in context?}
& Effect Estimation
& Effect Prediction
& Decision Support\\

\midrule

\rowcolor{gray!30}
\multicolumn{4}{c}{\textit{Counterfactual Causality}} \\

\makecell[l]{What would have happened\\if conditions had been\\different in context?}
& Imagination
& Counterfactual Reasoning
& Retrospective Analysis \\

\bottomrule
\end{tabular}
\caption{A taxonomy-driven overview of contextual causality with LLMs. The ``Core Question'' column specifies the causal question addressed by each category. The ``Capability'' column summarizes the causal ability required of LLMs for each category. The ``Tasks'' column lists representative tasks corresponding to each capability. The ``Practical Use'' column describes how each task can be employed in practical causality analysis.}
\label{tab:taxonomy}
\end{table*}

\section{Taxonomy of Contextual Causality} \label{sec:taxonomy}
In this section, we introduce our taxonomy of contextual causality, which consists of three categories: semantic, intervention, and counterfactual causality. These categories can be viewed as a progression from causal understanding to causal prediction and further to alternative-world causal reasoning. able \ref{tab:taxonomy} provides a detailed overview of the taxonomy. Prompt examples for representative tasks are provided in Appendix \ref{app:prompt-examples}.

\noindent \textbf{Semantic Causality}\quad This category concerns causal relations that are already expressed or implied in a given natural language context, focusing on what can be seen from the given text \cite{drury2022survey}. In this form of causality, causal relations are conveyed through explicit lexical markers or implicit semantic relations in the text, and can be identified from the given context alone. To understand semantic causality, LLMs need three major capabilities, including causal identification, causal explanation, and causal strength estimation.

Causal identification refers to the ability of LLMs to understand the complex causal structures expressed or implied in text. Tasks under this capability mainly include ECI \cite{mirza2014annotating}, causal judgment (CJ) \cite{du2022care}, causal attribution (CA) \cite{ho2022wikiwhy}, and causal detection and span extraction (CDS) \cite{ding2025multi}. ECI aims to determine whether there is a causal relationship between two events, thereby supporting causal graph construction in practice. CJ focuses on selecting the most plausible cause or effect of a target event from candidate options, supporting causal link refinement. That is, when multiple possible causal links are available, LLMs must determine which link is most likely to represent a valid causal relation. CA seeks to identify which event or factor is responsible for an observed outcome, corresponding to responsibility assignment in practical causality analysis. CDS focuses on detecting whether a text expresses causal meaning and, when applicable, extracting the textual spans corresponding to the cause and effect. In practice, it is mainly used to filter out texts lacking causal information and to provide cause-and-effect spans as textual evidence for the detected causal relation.

Causal explanation and causal strength estimation further extend semantic causality beyond the identification of causal structures. Causal explanation requires LLMs to explain the mechanism, process or logic through which a cause leads to its effect. The representative task under this capability is causal explanation generation (CEG) \cite{ho2022wikiwhy}, which aims to explain why a causal relation holds. In practice, CEG is mainly used to explain causal mechanisms, making causal relations more interpretable and useful for human analysis. Causal strength estimation requires LLMs to assess the degree to which one event causally influences or contributes to another. The representative task for this capability is causal strength assessment (CSA) \cite{romanou2023crab}, which goes beyond binary causal judgments by estimating the strength of the causal contribution between events. In practice, CSA is mainly used for quantifying causal influence, such as ranking causal factors or assigning strength scores to causal relations.

\noindent \textbf{Intervention Causality}\quad This category concerns the causal relationship between an introduced action and the outcome it produces in a given context. Here, ``intervention'' is used in a textual sense and differs from the classical causal inference notion of manipulating the underlying data-generating process. Unlike semantic causality, which focuses on understanding causal information already expressed or implied in the context, intervention causality asks what outcome would occur when an intervention is introduced into the given context, corresponding to the question ``What if I do...?'' Understanding intervention causality requires LLMs to perform effect estimation, i.e., estimating the likely outcomes of different actions in the given context. The representative task in this category is effect prediction (EP) \cite{li2026meter}, which concerns inferring the effect that a given action would have in a given contextual setting. In practice, EP supports decision-making by helping compare possible actions based on their expected consequences.

\noindent \textbf{Counterfactual Causality}\quad 
This category concerns whether and how the outcome would differ if a different action, event, or condition had occurred in a given context. 
Unlike intervention causality, counterfactual causality asks whether an observed outcome would have been different if a different action had been taken in the past, corresponding to the question ``What if I had done...?'' Understanding counterfactual causality requires LLMs to perform imagination, i.e., constructing an alternative scenario that differs from the observed facts and reasoning about how the outcome would change. The representative task under this category is counterfactual reasoning (CR) \cite{yu2023ifqa}, which aims to infer what would have happened under such hypothetical changes. In practice, CR supports retrospective analysis, such as reviewing clinical decisions, analyzing system failures, and examining whether an outcome could have been different under alternative conditions. Since CG has already been systematically surveyed in \cite{wang2024survey}, we do not discuss it further here.

\section{Contextual Causality with LLMs} \label{sec:method}
In this section, we review existing work in this area. We organize them into evaluation-oriented studies, including performance evaluation and benchmark construction for evaluating LLMs, and method-oriented studies, including training-free and training-based approaches. When a category lacks relevant studies, we omit that type.

\subsection{Performance Evaluation} 
Performance evaluation is essential for revealing the limitations of LLMs and identifying directions for future improvement.

\noindent \textbf{Semantic Causality}\quad Gao et al.~\cite{gao2023chatgpt} conduct the first systematic evaluation of various GPT-series models~\cite{brown2020language, achiam2023gpt} on ECI, CJ, and CEG. Their results show that these models suffer from severe causal hallucination in both ECI and CJ, tending to over-attribute causal relations even when no true causality exists. One possible explanation is that natural language contains a bias, as causal relations are often expressed with explicit cue words such as ``lead to'' and ``therefore'', while non-causal event pairs are rarely explicitly described. As a result, models trained on natural language may be more sensitive to causal relations. Nevertheless, GPT-series models remain relatively strong at explaining causal relations, as their generated explanations tend to be more complete and detailed. Zhao and Yan \cite{zhao2026generating} show that causal hallucination can be more severe in smaller LLMs ($\leq$1.5B parameters) on ECI. Their analysis of model responses suggests that these smaller models may mainly learn superficial answer templates. They further conduct a robustness analysis by introducing incorrect information as distracting evidence and find that model performance drops significantly, indicating sensitivity to misleading information. Meanwhile, Takayanagi et al. \cite{takayanagi2024chatgpt} find that causal hallucination also exists in CDS. 

Beyond causal hallucination, recent work~\cite{zhao2026mitigating} shows that different LLMs may exhibit distinct causal behaviors in ECI. For example, Qwen-series models~\cite{bai2023qwen} exhibit causal neglect, where they often refuse to infer a causal relation by claiming insufficient information, even when the provided context contains sufficient causal evidence. This behavior may be related to post-training data, such as instruction tuning, which can lead to more conservative responses.

Moreover, Yu et al.~\cite{yu2025causaleval} show that current LLMs, including DeepSeek, GPT, Claude, and others \cite{team2024gemma, grattafiori2024llama,liu2024deepseek, guo2025deepseek}, still have limited ability to handle contextual causality when the causal structure is complex, and the task goes beyond binary causal judgment. Although these models achieve high accuracy on short-context CJ, where the context usually contains only a few sentences and the causal structure is relatively simple, their performance drops sharply on CSA in CRAB. This performance drop suggests that current LLMs are strong on simple, short-context binary causal judgment, but still limited in complex causal analysis, such as cases where an event has multiple causes with different causal strengths under long contexts.

\noindent \textbf{Counterfactual Causality}\quad Kıcıman et al. \cite{kiciman2023causal} evaluate GPT-series models on CR and find that they can achieve strong performance, outperforming prior methods. Yu et al. \cite{yu2025causaleval} further evaluate a broad range of LLMs on CR. Their results show that advanced LLMs already perform very well on CR. However, the CRASS benchmark \cite{frohberg2022crass} used in these evaluations has a relatively simple structure, in which each example typically presents a single premise and asks what would have happened if that premise had not occurred. Therefore, strong performance on CRASS does not imply that LLMs have robust counterfactual reasoning abilities applicable to complex scenarios, as such scenarios often involve long contexts, multiple interacting events, and even unobserved factors.

\noindent \textbf{Limitations}\quad Existing performance evaluations of LLMs on contextual causality still provide an incomplete picture of models' causal abilities. First, current evaluations are highly imbalanced across different categories of contextual causality, mainly because benchmarks for intervention and counterfactual causality remain relatively scarce. Second, many studies evaluate only GPT-series models or a small set of LLMs, making their conclusions difficult to generalize across model families, especially since different models may exhibit distinct failure modes such as causal hallucination and neglect. Moreover, failure-mode analysis remains limited, even though different models may exhibit distinct failures such as causal hallucination and neglect. Third, existing evaluations often rely on final-answer accuracy, partly because fine-grained evaluation of intermediate reasoning processes is difficult to automate. This assessment makes it difficult to identify the true reasons behind LLMs' failures in contextual causality. Fourth, robustness analysis remains insufficient, with relatively few studies examining model behavior under misleading information. Finally, many existing benchmarks formulate contextual causal tasks as multiple-choice questions, which may allow models to rely on superficial shortcuts rather than genuinely understanding the underlying causal relations. Finally, many benchmarks remain simplified, so strong performance on them does not necessarily indicate that models have a deep understanding of contextual causality. Future evaluations should include more complex situations, such as long causal chains and events caused by multiple factors with various causal strengths.

\subsection{Benchmark Construction for LLMs} 
Benchmark construction is essential for evaluating the contextual causal understanding of LLMs.

\noindent \textbf{Semantic Causality}\quad Since existing QA datasets mostly focus on fact-seeking questions such as ``who'', ``what'', ``when'', and ``where'', Ho et al. \cite{ho2022wikiwhy} propose WikiWhy to support why-question answering. WikiWhy mainly covers CA and CEG. Similarly, although many benchmarks have been developed to evaluate LLMs' mathematical, coding, and general reasoning abilities, unified benchmarks for event relation extraction remain limited. To address this gap, Gong et al. \cite{gong2025eventrelbench} propose EventRelBench, a large-scale benchmark for evaluating LLMs on event relation extraction, where causal relation extraction, i.e., ECI, is included as one of the core tasks.

Motivated by the fact that human causal judgments are often graded rather than binary, Romanou et al. \cite{romanou2023crab} propose CRAB, which can be used to evaluate whether LLMs can estimate the strength of causal relations. In addition, to better evaluate whether LLMs can distinguish temporal relations from causal relations, Miliani et al. \cite{miliani2025explica} propose ExpliCa, a CJ benchmark. ExpliCITA \cite{bondielli2025llms} further extends ExpliCa to Italian. Considering data contamination, Chi et al.~\cite{chi2024unveiling} propose CausalProbe-2024, a CJ benchmark.

However, the above benchmarks are mainly constructed from news, Wikipedia, or human-constructed world-knowledge questions, which may still leave a gap relative to emerging real-world applications such as medical diagnostic records and legal case analysis. As LLM-based agents \cite{zhou2023webarena, yang2024swe, boiko2023autonomous} are increasingly used to assist humans in solving practical tasks, recent benchmarks have begun evaluating the causal attribution of LLMs across the execution trajectories of agentic systems. Zhang et al. \cite{zhang2025agent} propose Who\&When, which aims to identify which agent and which step cause task failures in LLM-based multi-agent systems. Similarly, Cemri et al. \cite{cemri2025multi} construct MAST-Data to analyze failure modes of these systems. However, these datasets still rely on human annotation, which is costly and time-consuming. To reduce this annotation burden, Zhang et al. \cite{zhang2025agentracer} propose AgenTracer, which automatically constructs failure-attribution data via counterfactual replay and program-fault injection. Recently, MP-Bench \cite{in2026rethinking} has also been proposed to rethink failure attribution in these systems. These benchmarks bring causal attribution closer to practical agent settings, but their data are still mainly generated from simple tasks, whereas real-world tasks, such as software iteration, involve multi-file changes and error-log diagnosis.

\noindent \textbf{Intervention Causality}\quad
Due to the lack of Portuguese causal benchmarks for evaluating LLMs, Lasheras and Pinheiro \cite{lasheras2025calquest} propose CaLQuest. The benchmark includes EP as a core component, but its examples are still mainly based on commonsense causal questions. 

\noindent \textbf{Counterfactual Causality}\quad
Since counterfactual reasoning is important for LLMs, Yu et al. \cite{yu2023ifqa} propose IfQA, a benchmark for open-domain question answering under counterfactual presuppositions. The benchmark is still mainly constructed from Wikipedia-based knowledge.

\noindent \textbf{Multi-level Contextual Causality}\quad
Although benchmarks have been proposed for different categories of contextual causality, they are often designed for isolated tasks and evaluated under separate contexts. This makes it difficult to assess whether LLMs can perform different types of contextual causality within the same scenario. To address this gap, Li et al. \cite{li2026meter} propose METER, a multi-level contextual causality benchmark that evaluates ECI, EP, and CR under a unified context. However, METER is mainly built by reorganizing existing benchmarks, so its underlying data sources still largely come from Wikipedia and news rather than real-world scenarios.

\noindent \textbf{Limitations}\quad Although recent studies have expanded benchmarks for evaluating LLMs on contextual causality, most benchmarks are not constructed from practical application scenarios such as clinical diagnosis records or legal case documents. As a result, these benchmarks cannot fully evaluate LLMs' ability to handle contextual causality in practical applications. Moreover, recent benchmarks based on the execution trajectories of agentic systems bring evaluation closer to practical scenarios, but their data are still mainly generated by simple tasks rather than long-horizon tasks that may take humans days or even months to complete, such as software iteration or scientific research. This may be because long-horizon benchmarks for agentic systems have recently emerged.

\subsection{Training-free}

Training-free methods are an important class of approaches for improving LLMs' performance, as they do not require substantial resources \cite{wei2022chain, wang2022self, du2023improving}.

\noindent \textbf{Semantic Causality}\quad 
Since LLMs may lack sufficient causal knowledge to directly identify event causality, Cai et al.~\cite{cai2025dr} propose Dr.ECI, a decomposed reasoning framework that incorporates causal structures into ECI. It decomposes ECI into multiple sub-tasks and uses multi-agent collaboration among a Causal Explorer, a Mediator Detector, and Direct and Indirect Reasoners. This framework enables LLMs to better identify both explicit and implicit causal relations. Similarly, Zou et al. \cite{zou2025mrbalance} propose MRBalance, which improves ECI through multi-agent debate with role assignment. Zeng et al. \cite{zeng2026zero} propose an ECI framework based on fuzzy aggregation of multisource evidence. However, these multi-agent approaches often incur substantial computational and time overhead. Zhao et al. \cite{zhao2026mitigating} propose a framework based on potential outcomes and actual causality theory that effectively mitigates causal bias in LLMs. However, these methods rely on the LLMs' ability to solve the decomposed sub-tasks. Recent work \cite{zhao2026generating} suggests that such frameworks may not generalize well to smaller models, and they require multiple calls to LLMs, leading to higher computational costs.

Beyond multi-agent collaboration, several works enhance LLMs' causal knowledge through external knowledge sources. Su et al. \cite{su2025enhancing} propose LKCER, which constructs a concept-level heterogeneous event graph and uses LLMs to enrich causal knowledge beyond the original event mentions. KnowQA \cite{wang2024document} reformulates document-level ECI as a binary question answering task. It leverages document-level event structures and designs single- and multi-turn causal QA strategies to identify causal relations. Wang et al. \cite{wang2024event} introduce a synthetic control method that uses retrieved similar event contexts to construct counterfactual controls for ECI. In addition, some studies also focus on other tasks. For example, Ashwani et al. \cite{ashwani2024cause} propose CARE-CA, a context-aware reasoning framework with counterfactual analysis. Chi et al. \cite{chi2024unveiling} propose G$^2$-Reasoner, which incorporates general knowledge and goal-oriented prompts into LLMs' causal reasoning process. Perak et al. \cite{perak2024incorporating} further explore whether retrieval-augmented generation (RAG) can help LLMs perform CJ more effectively. However, these methods depend on the coverage of external knowledge sources, which may limit their generalization to new scenarios, as each new scenario requires corresponding causal knowledge that is time-consuming to collect and difficult to make complete.

\noindent \textbf{Counterfactual Causality}\quad Since pretraining corpora of Code-LLMs contain many conditional structures, such as \texttt{if}, which naturally encode causal structures, Liu et al.~\cite{liu2023magic} explore whether Code-LLMs can exhibit stronger causal abilities than text-only LLMs. Their results show that Code-LLMs outperform text-only LLMs on CR. This suggests that exposure to training data with more explicit causal structures may help models acquire stronger causal reasoning abilities.

\noindent \textbf{Limitations}\quad Multi-agent-based methods may only be effective for specific models and often require multiple calls to LLMs, leading to higher computational cost. Methods based on external knowledge sources also rely heavily on the coverage of the provided knowledge. To generalize to different real-world scenarios, they would require a nearly complete causal knowledge base for each scenario, whose completeness is difficult to guarantee. These requirements make existing methods costly to apply and difficult to generalize. 

\subsection{Training-based}
Training remains the most effective way to inject knowledge into LLMs and improve their performance \cite{achiam2023gpt, guo2025deepseek}.

\noindent \textbf{Semantic Causality}\quad To mitigate causal hallucination in smaller models on ECI, Zhao and Yan \cite{zhao2026generating} empirically analyze the criteria for high-quality CoT traces. Based on these criteria, they design a pipeline to generate high-quality CoT traces and use them to fine-tune smaller models. Their results show that this method can effectively mitigate causal hallucination. In addition, Zhang et al. \cite{zhang2025agentracer} not only construct TracerTraj-2.5K to evaluate LLMs' causal attribution ability, but also train LLMs based on this dataset. Their results show that the trained LLMs achieve stronger causal attribution performance in execution trajectories of agentic systems. However, these methods are still task-specific.

\noindent \textbf{Limitations}\quad
Current fine-tuning datasets for contextual causality remain 
limited, with only a few available for training LLMs on specific tasks. These datasets are insufficient for comprehensively improving LLMs' contextual causal abilities. A unified dataset for improving LLMs on contextual causality is still lacking. One possible reason is that prior work has lacked a unified perspective on organizing contextual causality and has not sufficiently analyzed its role in addressing key challenges in current AI development, such as helping to build agents that can better solve real-world tasks through improved causal attribution. This may also limit exploration of reinforcement learning with verifiable rewards (RLVR), even though many contextual causality tasks are easily verifiable.

\begin{table}[!t]
\centering
\footnotesize
\setlength{\tabcolsep}{2pt}
\begin{tabular}{lcc}
\toprule
\textbf{Benchmark} & \textbf{Source} & \textbf{Focus} \\
\midrule
\rowcolor{gray!30}
\multicolumn{3}{c}{\textit{Semantic Causality}} \\
\multicolumn{3}{l}{\textit{Event Causal Identification}} \\
MAVEN-ERE \cite{wang2022maven}  & Wiki & Intrinsic \\
MECI \cite{lai2022meci} & Wiki & Intrinsic \\
ESL \cite{caselli2017event}  & News & Intrinsic \\
CTB \cite{mirza2014annotating}  & News & Intrinsic \\
BECauSE \cite{dunietz2017because} & News & Intrinsic \\
CaTeRS \cite{mostafazadeh2016caters}  & Stories & Intrinsic \\
METER \cite{li2026meter} & W\&N & Intrinsic \\
EventRelBench \cite{gong2025eventrelbench} & W\&N  & Intrinsic \\
\multicolumn{3}{l}{\textit{Causal Strength Assessment}} \\
CRAB \cite{romanou2023crab} & News & Intrinsic \\
\multicolumn{3}{l}{\textit{Causal judgment}} \\
COPA \cite{roemmele2011choice} & Human & Intrinsic \\
XCOPA \cite{ponti2020xcopa} & Human & Intrinsic \\
e-CARE \cite{du2022care} & Human & Intrinsic \\
ExpliCa \cite{miliani2025explica} & Human & Intrinsic \\
ExpliCITA \cite{bondielli2025llms} & Human & Intrinsic \\
MOCA \cite{nie2023moca} & Papers & Intrinsic \\
CausalProbe-2024 \cite{chi2024unveiling} & News & Intrinsic \\
\multicolumn{3}{l}{\textit{Causal Attribution}} \\
WikiWhy \cite{ho2022wikiwhy} & Wiki & Intrinsic \\
TellMeWhy \cite{lal2021tellmewhy} & Stories & Intrinsic \\
Who\&When \cite{zhang2025agent} & Agents & Intrinsic \\
TracerTraj-2.5K \cite{zhang2025agentracer} & Agents & Intrinsic \\
MP-Bench \cite{in2026rethinking} & Agents & Intrinsic \\
MAST-Data \citep{cemri2025multi} & Agents & Intrinsic \\
\multicolumn{3}{l}{\textit{Causal Explanation Generation}} \\
e-CARE \cite{du2022care} & Human & Intrinsic \\
WikiWhy \cite{ho2022wikiwhy} & Wiki & Intrinsic \\
ART \cite{bhagavatula2019abductive} & Stories & Intrinsic \\
\multicolumn{3}{l}{\textit{Causal Detection and Span Extraction}} \\
FinCausal2020 \cite{mariko2020financial} & News & Intrinsic \\
UniCausal \cite{tan2023unicausal} & News & Intrinsic \\
CausalTalk \cite{ding2025multi} & Reddit & Intrinsic \\
\rowcolor{gray!30}
\multicolumn{3}{c}{\textit{Intervention Causality}} \\
\multicolumn{3}{l}{\textit{Effect Prediction}} \\
METER \cite{li2026meter} & W\&N & Intrinsic \\
CaLQuest \cite{lasheras2025calquest} & Human & Intrinsic \\
\rowcolor{gray!30}
\multicolumn{3}{c}{\textit{Counterfactual Causality}} \\
\multicolumn{3}{l}{\textit{Counterfactual Reasoning}} \\
CRASS \cite{frohberg2022crass} & Human & Intrinsic \\
IfQA \cite{yu2023ifqa} & Wiki & Intrinsic \\
TimeTravel \cite{qin2019counterfactual} & Stories & Intrinsic \\
METER \cite{li2026meter} & W\&N & Intrinsic \\
\bottomrule
\end{tabular}
\caption{Representative benchmarks for contextual causality. ``Focus'' indicates whether a benchmark measures causality ability or how much the causality ability of current LLMs improves decision-making for real-world needs. ``Intrinsic'' denotes the former, while ``Extrinsic'' denotes the latter. ``Wiki'', ``W\&N'', ``Human'', and ``Agents'' denote Wikipedia-derived data, mixed Wikipedia and news sources, human-constructed based on world or commonsense knowledge, and agent-system traces or execution logs, respectively.}
\label{tab:causal_benchmarks}
\end{table}

\section{Benchmark Analysis} \label{sec:benchmark}

In this section, we analyze existing benchmarks. Representative benchmarks are summarized in Table~\ref{tab:causal_benchmarks}. Existing discussions \cite{yang2024critical} have examined whether these benchmarks can truly assess LLMs' contextual causality abilities or merely test their ability to retrieve knowledge. Our analysis focuses on the gap between current benchmarks and real-world demands.

From Table~\ref{tab:causal_benchmarks}, we can observe that they are mostly constructed from news, Wikipedia, stories, papers, social media, and human-written commonsense questions. However, strong performance on them may not indicate whether LLMs can perform contextual causal reasoning in real-world scenarios, where causal evidence is distributed across multiple lengthy, dynamic, and domain-specific sources, and causal analysis often requires reasoning beyond binary judgments over many-to-many causal relations with varying strengths. For example, in construction, practitioners need to analyze design documents, construction logs, and progress updates to identify the causes of delays and cost overruns, assess their relative impacts, and determine the main contributing factors.

In addition, recent benchmarks derived from agent-system traces or execution logs are closer to practical needs, as they can be used to evaluate LLMs' performance in identifying failure causes in agentic systems. However, these benchmarks are still built on simple tasks. In real-world applications, many tasks are long-horizon, often requiring days or even months of iterative work by humans, such as software development, where progress depends on requirements analysis, code modification, debugging, testing, and repeated revision, or scientific research, where hypotheses, experiments, and analyses evolve over multiple stages.

Table \ref{tab:causal_benchmarks} shows that existing benchmarks primarily evaluate specific contextual causal abilities of LLMs, such as causal judgment or causal attribution. In other words, most benchmarks adopt an intrinsic evaluation setting, where the goal is to test whether an LLM can identify or explain causal relations in a given context. However, there remains a lack of extrinsic evaluation measuring how much the contextual causal ability of current LLMs can improve the decision-making quality of AI systems or humans. However, there remains a lack of extrinsic evaluation measuring how much the contextual causal ability of current LLMs can improve the decision-making quality of AI systems or humans. This is important because the goal of evaluating and improving contextual causal reasoning is not merely to obtain higher performance on intrinsic benchmarks, but to enable AI systems or humans to make more reliable decisions.

\section{Future Directions} \label{sec:future}
\noindent \textbf{Real-world-oriented Benchmark Construction}\quad Future benchmarks for contextual causality should be designed to better align with real-world needs. First, future benchmarks should draw more from practical domains such as construction and healthcare, rather than relying on news and Wikipedia. In addition, future benchmarks could be constructed from agent-system execution trajectories in long-horizon tasks \cite{xiong2026autoresearchbench, thai2025swe, chen2025scienceagentbench}, where even humans may need days or months of iterative work to complete them. However, constructing such benchmarks first requires well-defined long-horizon benchmarks for agents, so that meaningful execution trajectories can be collected for causal attribution evaluation. Second, future benchmarks should include extrinsic settings that assess whether LLMs can leverage contextual causality to improve the decision-making of AI systems or humans, for example, by evaluating whether LLM-generated causal graphs improve the decision quality of agents.

\noindent \textbf{More Comprehensive and Fine-grained Performance Evaluation}\quad 
Future evaluations should provide a more complete picture of LLMs' contextual causal abilities. First, they should cover a broader range of tasks, especially intervention and counterfactual causality, which remain less explored due to limited benchmarks. Second, they should include more diverse model families, since different LLMs may exhibit different causal failure modes, such as causal hallucination or neglect. Third, future evaluations should move beyond final-answer accuracy toward fine-grained evaluation \cite{zheng2025survey, li2025generation, chern2023factool, min2023factscore}. Recent work on process reward models (PRMs) provides useful methodological guidance for evaluating intermediate reasoning steps through step-level supervision and learned verifiers or reward models \cite{zheng2025survey}. Fine-grained evaluation methods such as FActScore and FacTool further illustrate how model outputs can be decomposed into smaller units and verified individually \cite{chern2023factool, min2023factscore}. For contextual causality, merely judging the final answer is insufficient, as models may obtain correct answers through memorized knowledge or task-specific shortcuts rather than genuine understanding of contextual causality. Therefore, future work should develop automated methods for evaluating the intermediate processes underlying LLM responses.

\noindent \textbf{Post-training Data Construction}\quad Since prior studies have suggested that LLMs' causal hallucination and neglect may be related to limitations in training data, and existing data mainly focus on individual settings, such as ECI or CA in agentic systems, future work should construct more comprehensive post-training data for contextual causality \cite{xu2025kodcode, xu2025toucan}. A promising direction is to build unified datasets covering representative tasks across semantic, intervention, and counterfactual causality. For each sample in the dataset, the response should also include sufficiently detailed intermediate steps \cite{ye2025limo, chen2026towards,li2025system}, rather than only the final answer, so that LLMs can learn the underlying causal reasoning process and comprehensively improve their contextual causal abilities. In addition, when constructing fine-tuning data, future work should also consider effective data mixture \cite{li2025small}, quality filtering \cite{zhang2026quest}, and reducing the distribution gap between the CoT data and target models \cite{zhao2026generating}.

\noindent \textbf{Causal Hallucination Attribution}\quad Causal hallucination is an important manifestation of the limited contextual causal abilities of current LLMs, yet its causes remain insufficiently studied. Future work should systematically investigate causal hallucination, drawing on broader studies of LLM hallucination \cite{dziri2022origin, ji2023survey, kalai2025language}. Such attribution can help identify when and why LLMs produce incorrect causal relations and provide guidance for developing targeted mitigation methods.

\noindent \textbf{Need for More Attention to Intervention and Counterfactual Causality}\quad Previous research has mainly focused on semantic causality, partly because the community has been concerned with improving LLMs' language understanding, knowledge use, and basic reasoning abilities. However, as LLMs are increasingly used to autonomously solve real-world tasks \cite{wei2026agentic, gupta2025llms}, decision-making becomes more central. Intervention and counterfactual causality are closely tied to this goal, because they help LLMs estimate the effects of possible actions and reflect on how past decisions could be improved. Therefore, future research should place more emphasis on intervention and counterfactual causality, including benchmark construction, performance evaluation, and method development.

\section{Conclusion}
In this survey, we present an overview of contextual causality with LLMs. We first propose a taxonomy of contextual causality motivated by the needs of causal analysis. This taxonomy organizes contextual causality into semantic, intervention, and counterfactual causality, and further characterizes each category by its core causal question, required capability, representative tasks, and practical role in causality analysis. We then summarize and analyze existing studies and benchmarks. Our analysis reveals several key limitations in current research, including limited, coarse-grained performance evaluation, insufficient post-training data, limited attention to interventions and counterfactual causality, and a gap between current benchmarks and real-world needs. Lastly, we discuss several promising directions for future research. We hope this survey can provide a roadmap for this field, highlight its importance, and promote future research toward solving real-world problems.

\section*{Limitations}
This survey focuses on contextual causality in text-only scenarios. However, the proposed taxonomy can also be extended to multimodal settings. For example, in an embodied scenario, foundation models need to identify causal relations from visual observations, language instructions, and action histories, predict the effects of possible interventions, reason about counterfactual outcomes under alternative actions, and finally make reliable decisions in the environment. This suggests that contextual causality is also important for multimodal foundation models, where contextual causality may involve images, videos, and sensor observations in addition to text. In future work, we plan to extend this survey from LLMs to multimodal foundation models, including vision-language models (VLMs), vision-language-action models (VLAs), and vision-language-navigation models (VLNs). In addition, our taxonomy is grounded in classical causal theory and primarily serves as a unified framework for analyzing contextual causality in LLMs rather than extending causal theory itself. Strengthening the theoretical foundation of the taxonomy remains an important direction for future work.

\section*{Acknowledgments}
We gratefully acknowledge the support of the Natural Sciences and Engineering Research Council of Canada (NSERC) under Grants ALLRP 585937-23 and RGPIN-2025-05097, and Mitacs under Grant IT35587.

\bibliography{custom}

@article{lake2017building,
  title={Building machines that learn and think like people},
  author={Lake, Brenden M and Ullman, Tomer D and Tenenbaum, Joshua B and Gershman, Samuel J},
  journal={Behavioral and brain sciences},
  volume={40},
  pages={e253},
  year={2017},
  publisher={Cambridge University Press}
}

@article{drury2022survey,
  title={A survey of the extraction and applications of causal relations},
  author={Drury, Brett and Oliveira, Hugo Gon{\c{c}}alo and de Andrade Lopes, Alneu},
  journal={Natural Language Engineering},
  volume={28},
  number={3},
  pages={361--400},
  year={2022},
  publisher={Cambridge University Press}
}

@inproceedings{wang2024causalbench,
  title={Causalbench: A comprehensive benchmark for evaluating causal reasoning capabilities of large language models},
  author={Wang, Zeyu},
  booktitle={Proceedings of the 10th SIGHAN Workshop on Chinese Language Processing (SIGHAN-10)},
  pages={143--151},
  year={2024}
}

@article{yang2022survey,
  title={A survey on extraction of causal relations from natural language text},
  author={Yang, Jie and Han, Soyeon Caren and Poon, Josiah},
  journal={Knowledge and Information Systems},
  volume={64},
  number={5},
  pages={1161--1186},
  year={2022},
  publisher={Springer}
}

@article{richens2020improving,
  title={Improving the accuracy of medical diagnosis with causal machine learning},
  author={Richens, Jonathan G and Lee, Ciar{\'a}n M and Johri, Saurabh},
  journal={Nature communications},
  volume={11},
  number={1},
  pages={3923},
  year={2020},
  publisher={Nature Publishing Group UK London}
}

@article{wu2024causal,
  title={Causal inference in the medical domain: a survey: X. Wu et al.},
  author={Wu, Xing and Peng, Shaoqi and Li, Jingwen and Zhang, Jian and Sun, Qun and Li, Weimin and Qian, Quan and Liu, Yue and Guo, Yike},
  journal={Applied Intelligence},
  volume={54},
  number={6},
  pages={4911--4934},
  year={2024},
  publisher={Springer}
}

@book{halpern2016actual,
  title={Actual causality},
  author={Halpern, Joseph Y},
  year={2016},
  publisher={MiT Press}
}

@article{wan2024large,
  title={Large language models for causal discovery: Current landscape and future directions},
  author={Wan, Guangya and Lu, Yunsheng and Wu, Yuqi and Hu, Mengxuan and Li, Sheng},
  journal={arXiv preprint arXiv:2402.11068},
  year={2024}
}

@article{liu2025large,
  title={Large language models and causal inference in collaboration: A comprehensive survey},
  author={Liu, Xiaoyu and Xu, Paiheng and Wu, Junda and Yuan, Jiaxin and Yang, Yifan and Zhou, Yuhang and Liu, Fuxiao and Guan, Tianrui and Wang, Haoliang and Yu, Tong and others},
  journal={Findings of the Association for Computational Linguistics: NAACL 2025},
  pages={7668--7684},
  year={2025}
}

@article{ma2025causal,
  title={Causal inference with large language model: A survey},
  author={Ma, Jing},
  journal={Findings of the Association for Computational Linguistics: NAACL 2025},
  pages={5886--5898},
  year={2025}
}

@inproceedings{yu2025causaleval,
  title={Causaleval: Towards better causal reasoning in language models},
  author={Yu, Longxuan and Chen, Delin and Xiong, Siheng and Wu, Qingyang and Li, Dawei and Chen, Zhikai and Liu, Xiaoze and Pan, Liangming},
  booktitle={Proceedings of the 2025 Conference of the Nations of the Americas Chapter of the Association for Computational Linguistics: Human Language Technologies (Volume 1: Long Papers)},
  pages={12512--12540},
  year={2025}
}

@article{zhou2025emerging,
  title={Emerging synergies in causality and deep generative models: A survey},
  author={Zhou, Guanglin and Xie, Shaoan and Hao, Guang-Yuan and Chen, Shiming and Huang, Biwei and Xu, Xiwei and Wang, Chen and Zhu, Liming and Yao, Lina and Zhang, Kun},
  journal={IEEE Transactions on Artificial Intelligence},
  year={2025},
  publisher={IEEE}
}

@article{bazgir2025causal,
  title={Causal mas: A survey of large language model architectures for discovery and effect estimation},
  author={Bazgir, Adib and Habibdoust, Amir and Zhang, Yuwen and Song, Xing},
  journal={arXiv preprint arXiv:2509.00987},
  year={2025}
}

@inproceedings{li2025survey,
  title={A survey on enhancing causal reasoning ability of large language models},
  author={Li, Xin and Cai, Zhuo and Wang, Shoujin and Yu, Kun and Chen, Fang},
  booktitle={Pacific-Asia Conference on Knowledge Discovery and Data Mining},
  pages={399--416},
  year={2025},
  organization={Springer}
}

@inproceedings{cui2024odyssey,
  title={The odyssey of commonsense causality: From foundational benchmarks to cutting-edge reasoning},
  author={Cui, Shaobo and Jin, Zhijing and Sch{\"o}lkopf, Bernhard and Faltings, Boi},
  booktitle={Proceedings of the 2024 Conference on Empirical Methods in Natural Language Processing},
  pages={16722--16763},
  year={2024}
}

@inproceedings{cui2025uncertainty,
  title={Uncertainty in causality: A new frontier},
  author={Cui, Shaobo and Mouchel, Luca and Faltings, Boi},
  booktitle={Proceedings of the 63rd Annual Meeting of the Association for Computational Linguistics (Volume 1: Long Papers)},
  pages={8022--8044},
  year={2025}
}

@article{yang2024critical,
  title={A critical review of causal reasoning benchmarks for large language models},
  author={Yang, Linying and Shirvaikar, Vik and Clivio, Oscar and Falck, Fabian},
  journal={arXiv preprint arXiv:2407.08029},
  year={2024}
}

@inproceedings{wang2024survey,
  title={A survey on natural language counterfactual generation},
  author={Wang, Yongjie and Qiu, Xiaoqi and Yue, Yu and Guo, Xu and Zeng, Zhiwei and Feng, Yuhong and Shen, Zhiqi},
  booktitle={Findings of the Association for Computational Linguistics: EMNLP 2024},
  pages={4798--4818},
  year={2024}
}

@article{cheng2025survey,
  title={A Survey of Event Causality Identification: Taxonomy, Challenges, Assessment, and Prospects},
  author={Cheng, Qing and Zeng, Zefan and Hu, Xingchen and Si, Yuehang and Liu, Zhong},
  journal={ACM Computing Surveys},
  volume={58},
  number={3},
  pages={1--37},
  year={2025},
  publisher={ACM New York, NY}
}

@inproceedings{gao2023chatgpt,
  title={Is chatgpt a good causal reasoner? a comprehensive evaluation},
  author={Gao, Jinglong and Ding, Xiao and Qin, Bing and Liu, Ting},
  booktitle={Findings of the association for computational linguistics: EMNLP 2023},
  pages={11111--11126},
  year={2023}
}

@article{wei2022chain,
  title={Chain-of-thought prompting elicits reasoning in large language models},
  author={Wei, Jason and Wang, Xuezhi and Schuurmans, Dale and Bosma, Maarten and Xia, Fei and Chi, Ed and Le, Quoc V and Zhou, Denny and others},
  journal={Advances in neural information processing systems},
  volume={35},
  pages={24824--24837},
  year={2022}
}

@article{brown2020language,
  title={Language models are few-shot learners},
  author={Brown, Tom and Mann, Benjamin and Ryder, Nick and Subbiah, Melanie and Kaplan, Jared D and Dhariwal, Prafulla and Neelakantan, Arvind and Shyam, Pranav and Sastry, Girish and Askell, Amanda and others},
  journal={Advances in neural information processing systems},
  volume={33},
  pages={1877--1901},
  year={2020}
}

@article{achiam2023gpt,
  title={Gpt-4 technical report},
  author={Achiam, Josh and Adler, Steven and Agarwal, Sandhini and Ahmad, Lama and Akkaya, Ilge and Aleman, Florencia Leoni and Almeida, Diogo and Altenschmidt, Janko and Altman, Sam and Anadkat, Shyamal and others},
  journal={arXiv preprint arXiv:2303.08774},
  year={2023}
}

@article{zhao2026generating,
  title={Generating Effective CoT Traces for Mitigating Causal Hallucination},
  author={Zhao, Yiheng and Yan, Jun},
  journal={arXiv preprint arXiv:2604.12748},
  year={2026}
}

@inproceedings{zhao2026mitigating,
  title={Mitigating Causal Bias in LLMs via Potential Outcomes Framework and Actual Causality Theory},
  author={Zhao, Yiheng and Li, Yuanliang and Savant, Shreya and Yan, Jun},
  booktitle={Findings of the Association for Computational Linguistics: EACL 2026},
  pages={4212--4222},
  year={2026}
}

@article{bai2023qwen,
  title={Qwen technical report},
  author={Bai, Jinze and Bai, Shuai and Chu, Yunfei and Cui, Zeyu and Dang, Kai and Deng, Xiaodong and Fan, Yang and Ge, Wenbin and Han, Yu and Huang, Fei and others},
  journal={arXiv preprint arXiv:2309.16609},
  year={2023}
}

@inproceedings{takayanagi2024chatgpt,
  title={Is chatgpt the future of causal text mining? a comprehensive evaluation and analysis},
  author={Takayanagi, Takehiro and Suzuki, Masahiro and Kobayashi, Ryotaro and Sakaji, Hiroki and Izumi, Kiyoshi},
  booktitle={2024 IEEE International Conference on Big Data (BigData)},
  pages={6651--6660},
  year={2024},
  organization={IEEE}
}

@article{team2024gemma,
  title={Gemma 2: Improving open language models at a practical size},
  author={Team, Gemma and Riviere, Morgane and Pathak, Shreya and Sessa, Pier Giuseppe and Hardin, Cassidy and Bhupatiraju, Surya and Hussenot, L{\'e}onard and Mesnard, Thomas and Shahriari, Bobak and Ram{\'e}, Alexandre and others},
  journal={arXiv preprint arXiv:2408.00118},
  year={2024}
}

@article{grattafiori2024llama,
  title={The llama 3 herd of models},
  author={Grattafiori, Aaron and Dubey, Abhimanyu and Jauhri, Abhinav and Pandey, Abhinav and Kadian, Abhishek and Al-Dahle, Ahmad and Letman, Aiesha and Mathur, Akhil and Schelten, Alan and Vaughan, Alex and others},
  journal={arXiv preprint arXiv:2407.21783},
  year={2024}
}

@article{liu2024deepseek,
  title={Deepseek-v3 technical report},
  author={Liu, Aixin and Feng, Bei and Xue, Bing and Wang, Bingxuan and Wu, Bochao and Lu, Chengda and Zhao, Chenggang and Deng, Chengqi and Zhang, Chenyu and Ruan, Chong and others},
  journal={arXiv preprint arXiv:2412.19437},
  year={2024}
}

@article{guo2025deepseek,
  title={Deepseek-r1: Incentivizing reasoning capability in llms via reinforcement learning},
  author={Guo, Daya and Yang, Dejian and Zhang, Haowei and Song, Junxiao and Wang, Peiyi and Zhu, Qihao and Xu, Runxin and Zhang, Ruoyu and Ma, Shirong and Bi, Xiao and others},
  journal={arXiv preprint arXiv:2501.12948},
  year={2025}
}

@article{kiciman2023causal,
  title={Causal reasoning and large language models: Opening a new frontier for causality},
  author={Kiciman, Emre and Ness, Robert and Sharma, Amit and Tan, Chenhao},
  journal={Transactions on Machine Learning Research},
  year={2023}
}

@article{ho2022wikiwhy,
  title={Wikiwhy: Answering and explaining cause-and-effect questions},
  author={Ho, Matthew and Sharma, Aditya and Chang, Justin and Saxon, Michael and Levy, Sharon and Lu, Yujie and Wang, William Yang},
  journal={arXiv preprint arXiv:2210.12152},
  year={2022}
}

@inproceedings{gong2025eventrelbench,
  title={EventRelBench: A Comprehensive Benchmark for Evaluating Event Relation Understanding in Large Language Models},
  author={Gong, Jie and Zheng, Biaoshuai and Hu, Qiwang},
  booktitle={Findings of the Association for Computational Linguistics: EMNLP 2025},
  pages={9084--9099},
  year={2025}
}

@inproceedings{romanou2023crab,
  title={Crab: Assessing the strength of causal relationships between real-world events},
  author={Romanou, Angelika and Montariol, Syrielle and Paul, Debjit and Laugier, Leo and Aberer, Karl and Bosselut, Antoine},
  booktitle={Proceedings of the 2023 Conference on Empirical Methods in Natural Language Processing},
  pages={15198--15216},
  year={2023}
}

@inproceedings{miliani2025explica,
  title={ExpliCa: Evaluating explicit causal reasoning in large language models},
  author={Miliani, Martina and Auriemma, Serena and Bondielli, Alessandro and Chersoni, Emmanuele and Passaro, Lucia and Sucameli, Irene and Lenci, Alessandro},
  booktitle={Findings of the Association for Computational Linguistics: ACL 2025},
  pages={17335--17355},
  year={2025}
}

@inproceedings{bondielli2025llms,
  title={LLMs Struggle on Explicit Causality in Italian},
  author={Bondielli, Alessandro and Miliani, Martina and Paglione, Luca and Auriemma, Serena and Passaro, Lucia C and Lenci, Alessandro},
  booktitle={Proceedings of the Eleventh Italian Conference on Computational Linguistics (CLiC-it 2025)},
  pages={83--94},
  year={2025}
}

@article{chi2024unveiling,
  title={Unveiling causal reasoning in large language models: Reality or mirage?},
  author={Chi, Haoang and Li, He and Yang, Wenjing and Liu, Feng and Lan, Long and Ren, Xiaoguang and Liu, Tongliang and Han, Bo},
  journal={Advances in Neural Information Processing Systems},
  volume={37},
  pages={96640--96670},
  year={2024}
}

@article{zhang2025agent,
  title={Which agent causes task failures and when? on automated failure attribution of llm multi-agent systems},
  author={Zhang, Shaokun and Yin, Ming and Zhang, Jieyu and Liu, Jiale and Han, Zhiguang and Zhang, Jingyang and Li, Beibin and Wang, Chi and Wang, Huazheng and Chen, Yiran and others},
  journal={arXiv preprint arXiv:2505.00212},
  year={2025}
}

@article{cemri2025multi,
  title={Why do multi-agent llm systems fail?},
  author={Cemri, Mert and Pan, Melissa Z and Yang, Shuyi and Agrawal, Lakshya A and Chopra, Bhavya and Tiwari, Rishabh and Keutzer, Kurt and Parameswaran, Aditya and Klein, Dan and Ramchandran, Kannan and others},
  journal={arXiv preprint arXiv:2503.13657},
  year={2025}
}

@article{zhang2025agentracer,
  title={AgenTracer: Who Is Inducing Failure in the LLM Agentic Systems?},
  author={Zhang, Guibin and Wang, Junhao and Chen, Junjie and Zhou, Wangchunshu and Wang, Kun and Yan, Shuicheng},
  journal={arXiv preprint arXiv:2509.03312},
  year={2025}
}

@article{zhou2023webarena,
  title={Webarena: A realistic web environment for building autonomous agents},
  author={Zhou, Shuyan and Xu, Frank F and Zhu, Hao and Zhou, Xuhui and Lo, Robert and Sridhar, Abishek and Cheng, Xianyi and Ou, Tianyue and Bisk, Yonatan and Fried, Daniel and others},
  journal={arXiv preprint arXiv:2307.13854},
  year={2023}
}

@article{yang2024swe,
  title={Swe-agent: Agent-computer interfaces enable automated software engineering},
  author={Yang, John and Jimenez, Carlos E and Wettig, Alexander and Lieret, Kilian and Yao, Shunyu and Narasimhan, Karthik and Press, Ofir},
  journal={Advances in Neural Information Processing Systems},
  volume={37},
  pages={50528--50652},
  year={2024}
}

@article{boiko2023autonomous,
  title={Autonomous chemical research with large language models},
  author={Boiko, Daniil A and MacKnight, Robert and Kline, Ben and Gomes, Gabe},
  journal={Nature},
  volume={624},
  number={7992},
  pages={570--578},
  year={2023},
  publisher={Nature Publishing Group UK London}
}

@inproceedings{lasheras2025calquest,
  title={Calquest. pt: Towards the collection and evaluation of natural causal ladder questions in portuguese for ai agents},
  author={Lasheras, Uriel Anderson and Pinheiro, Vl{\'a}dia},
  booktitle={Proceedings of the First Workshop on Language Models for Low-Resource Languages},
  pages={325--343},
  year={2025}
}

@inproceedings{yu2023ifqa,
  title={Ifqa: A dataset for open-domain question answering under counterfactual presuppositions},
  author={Yu, Wenhao and Jiang, Meng and Clark, Peter and Sabharwal, Ashish},
  booktitle={Proceedings of the 2023 Conference on Empirical Methods in Natural Language Processing},
  pages={8276--8288},
  year={2023}
}

@article{li2026meter,
  title={METER: Evaluating Multi-Level Contextual Causal Reasoning in Large Language Models},
  author={Li, Pengfeng and Huang, Chen and Hao, Chaoqun and Chen, Hongyao and Wei, Xiao-Yong and Lei, Wenqiang and Ng, See-Kiong},
  journal={arXiv preprint arXiv:2604.11502},
  year={2026}
}

@article{wang2022self,
  title={Self-consistency improves chain of thought reasoning in language models},
  author={Wang, Xuezhi and Wei, Jason and Schuurmans, Dale and Le, Quoc and Chi, Ed and Narang, Sharan and Chowdhery, Aakanksha and Zhou, Denny},
  journal={arXiv preprint arXiv:2203.11171},
  year={2022}
}

@article{du2023improving,
  title={Improving factuality and reasoning in language models through multiagent debate},
  author={Du, Yilun and Li, Shuang and Torralba, Antonio and Tenenbaum, Joshua B and Mordatch, Igor},
  journal={arXiv preprint arXiv:2305.14325},
  year={2023}
}

@inproceedings{cai2025dr,
  title={Dr. ECI: Infusing large language models with causal knowledge for decomposed reasoning in event causality identification},
  author={Cai, Ruichu and Yu, Shengyin and Zhang, Jiahao and Chen, Wei and Xu, Boyan and Zhang, Keli},
  booktitle={Proceedings of the 31st International Conference on Computational Linguistics},
  pages={9346--9375},
  year={2025}
}

@article{zou2025mrbalance,
  title={MRBalance: A Framework for Enhancing Event Causality Identification in Multi-Agent Debates via Role Assignment},
  author={Zou, Xiang and Li, Xuanhong and Hu, Po and Dong, Ming},
  journal={Knowledge-Based Systems},
  pages={114470},
  year={2025},
  publisher={Elsevier}
}

@article{zeng2026zero,
  title={Zero-Shot Event Causality Identification via Multisource Evidence Fuzzy Aggregation With Large Language Models},
  author={Zeng, Zefan and Cheng, Qing and Hu, Xingchen and Li, Wentao and Ding, Weiping and Liu, Zhong},
  journal={IEEE Transactions on Fuzzy Systems},
  year={2026},
  publisher={IEEE}
}

@inproceedings{su2025enhancing,
  title={Enhancing event causality identification with LLM knowledge and concept-level event relations},
  author={Su, Ya and Zhang, Hu and Zhang, Guangjun and Wang, Yujie and Fan, Yue and Li, Ru and Wang, Yuanlong},
  booktitle={Proceedings of the 31st International Conference on Computational Linguistics},
  pages={7403--7414},
  year={2025}
}

@inproceedings{wang2024document,
  title={Document-level causal relation extraction with knowledge-guided binary question answering},
  author={Wang, Zimu and Xia, Lei and Wang, Wei and Du, Xinya},
  booktitle={Findings of the Association for Computational Linguistics: EMNLP 2024},
  pages={16944--16955},
  year={2024}
}

@inproceedings{wang2024event,
  title={Event causality identification with synthetic control},
  author={Wang, Haoyu and Liu, Fengze and Zhang, Jiayao and Roth, Dan and Richardson, Kyle},
  booktitle={Proceedings of the 2024 Conference on Empirical Methods in Natural Language Processing},
  pages={1725--1737},
  year={2024}
}

@inproceedings{ashwani2024cause,
  title={Cause and effect: Can large language models truly understand causality?},
  author={Ashwani, Swagata and Hegde, Kshiteesh and Mannuru, Nishith Reddy and Sengar, Dushyant Singh and Jindal, Mayank and Kathala, Krishna Chaitanya Rao and Banga, Dishant and Jain, Vinija and Chadha, Aman},
  booktitle={Proceedings of the AAAI Symposium Series},
  volume={4},
  number={1},
  pages={2--9},
  year={2024}
}

@inproceedings{perak2024incorporating,
  title={Incorporating dialect understanding into LLM using RAG and prompt engineering techniques for causal commonsense reasoning},
  author={Perak, Benedikt and Beliga, Slobodan and Me{\v{s}}trovi{\'c}, Ana},
  booktitle={Proceedings of the Eleventh Workshop on NLP for Similar Languages, Varieties, and Dialects (VarDial 2024)},
  pages={220--229},
  year={2024}
}

@inproceedings{liu2023magic,
  title={The magic of IF: Investigating causal reasoning abilities in large language models of code},
  author={Liu, Xiao and Yin, Da and Zhang, Chen and Feng, Yansong and Zhao, Dongyan},
  booktitle={Findings of the Association for Computational Linguistics: ACL 2023},
  pages={9009--9022},
  year={2023}
}

@inproceedings{wang2022maven,
  title={Maven-ere: A unified large-scale dataset for event coreference, temporal, causal, and subevent relation extraction},
  author={Wang, Xiaozhi and Chen, Yulin and Ding, Ning and Peng, Hao and Wang, Zimu and Lin, Yankai and Han, Xu and Hou, Lei and Li, Juanzi and Liu, Zhiyuan and others},
  booktitle={Proceedings of the 2022 Conference on Empirical Methods in Natural Language Processing},
  pages={926--941},
  year={2022}
}

@inproceedings{lai2022meci,
  title={MECI: A multilingual dataset for event causality identification},
  author={Lai, Viet Dac and Veyseh, Amir Pouran Ben and Van Nguyen, Minh and Dernoncourt, Franck and Nguyen, Thien Huu},
  booktitle={Proceedings of the 29th international conference on computational linguistics},
  pages={2346--2356},
  year={2022}
}

@inproceedings{caselli2017event,
  title={The event storyline corpus: A new benchmark for causal and temporal relation extraction},
  author={Caselli, Tommaso and Vossen, Piek},
  booktitle={Proceedings of the Events and Stories in the News Workshop},
  pages={77--86},
  year={2017}
}

@inproceedings{mirza2014annotating,
  title={Annotating causality in the TempEval-3 corpus},
  author={Mirza, Paramita and Sprugnoli, Rachele and Tonelli, Sara and Speranza, Manuela},
  booktitle={Proceedings of the EACL 2014 workshop on computational approaches to causality in language (CAtoCL)},
  pages={10--19},
  year={2014}
}

@inproceedings{dunietz2017because,
  title={The BECauSE corpus 2.0: Annotating causality and overlapping relations},
  author={Dunietz, Jesse and Levin, Lori and Carbonell, Jaime G},
  booktitle={Proceedings of the 11th linguistic annotation workshop},
  pages={95--104},
  year={2017}
}

@inproceedings{mostafazadeh2016caters,
  title={CaTeRS: Causal and temporal relation scheme for semantic annotation of event structures},
  author={Mostafazadeh, Nasrin and Grealish, Alyson and Chambers, Nathanael and Allen, James and Vanderwende, Lucy},
  booktitle={Proceedings of the fourth workshop on events},
  pages={51--61},
  year={2016}
}

@inproceedings{roemmele2011choice,
  title={Choice of Plausible Alternatives: An Evaluation of Commonsense Causal Reasoning.},
  author={Roemmele, Melissa and Bejan, Cosmin Adrian and Gordon, Andrew S},
  booktitle={AAAI spring symposium: logical formalizations of commonsense reasoning},
  pages={90--95},
  year={2011}
}

@inproceedings{ponti2020xcopa,
  title={XCOPA: A multilingual dataset for causal commonsense reasoning},
  author={Ponti, Edoardo Maria and Glava{\v{s}}, Goran and Majewska, Olga and Liu, Qianchu and Vuli{\'c}, Ivan and Korhonen, Anna},
  booktitle={Proceedings of the 2020 Conference on Empirical Methods in Natural Language Processing (EMNLP)},
  pages={2362--2376},
  year={2020}
}

@inproceedings{du2022care,
  title={e-CARE: a new dataset for exploring explainable causal reasoning},
  author={Du, Li and Ding, Xiao and Xiong, Kai and Liu, Ting and Qin, Bing},
  booktitle={Proceedings of the 60th Annual Meeting of the Association for Computational Linguistics (Volume 1: Long Papers)},
  pages={432--446},
  year={2022}
}

@article{nie2023moca,
  title={Moca: Measuring human-language model alignment on causal and moral judgment tasks},
  author={Nie, Allen and Zhang, Yuhui and Amdekar, Atharva Shailesh and Piech, Chris and Hashimoto, Tatsunori B and Gerstenberg, Tobias},
  journal={Advances in Neural Information Processing Systems},
  volume={36},
  pages={78360--78393},
  year={2023}
}

@inproceedings{lal2021tellmewhy,
  title={TellMeWhy: A dataset for answering why-questions in narratives},
  author={Lal, Yash Kumar and Chambers, Nathanael and Mooney, Raymond and Balasubramanian, Niranjan},
  booktitle={Findings of the Association for Computational Linguistics: ACL-IJCNLP 2021},
  pages={596--610},
  year={2021}
}

@article{in2026rethinking,
  title={Rethinking Failure Attribution in Multi-Agent Systems: A Multi-Perspective Benchmark and Evaluation},
  author={In, Yeonjun and Tanjim, Mehrab and Subramanian, Jayakumar and Kim, Sungchul and Bhattacharya, Uttaran and Kim, Wonjoong and Park, Sangwu and Sarkhel, Somdeb and Park, Chanyoung},
  journal={arXiv preprint arXiv:2603.25001},
  year={2026}
}

@article{bhagavatula2019abductive,
  title={Abductive commonsense reasoning},
  author={Bhagavatula, Chandra and Bras, Ronan Le and Malaviya, Chaitanya and Sakaguchi, Keisuke and Holtzman, Ari and Rashkin, Hannah and Downey, Doug and Yih, Scott Wen-tau and Choi, Yejin},
  journal={arXiv preprint arXiv:1908.05739},
  year={2019}
}

@inproceedings{mariko2020financial,
  title={The financial document causality detection shared task (FinCausal 2020)},
  author={Mariko, Dominique and Abi-Akl, Hanna and Labidurie, Estelle and Durfort, Stephane and De Mazancourt, Hugues and El-Haj, Mahmoud},
  booktitle={Proceedings of the 1st Joint Workshop on Financial Narrative Processing and MultiLing Financial Summarisation},
  pages={23--32},
  year={2020}
}

@inproceedings{tan2023unicausal,
  title={Unicausal: Unified benchmark and repository for causal text mining},
  author={Tan, Fiona Anting and Zuo, Xinyu and Ng, See-Kiong},
  booktitle={International Conference on Big Data Analytics and Knowledge Discovery},
  pages={248--262},
  year={2023},
  organization={Springer}
}

@inproceedings{ding2025multi,
  title={A multi-level benchmark for causal language understanding in social media discourse},
  author={Ding, Xiaohan and Ping, Kaike and {\c{C}}ar{\i}k, Buse and Rho, Eugenia},
  booktitle={Proceedings of the 2025 Conference on Empirical Methods in Natural Language Processing},
  pages={28764--28778},
  year={2025}
}

@inproceedings{frohberg2022crass,
  title={CRASS: A novel data set and benchmark to test counterfactual reasoning of large language models},
  author={Frohberg, J{\"o}rg and Binder, Frank},
  booktitle={Proceedings of the Thirteenth Language Resources and Evaluation Conference},
  pages={2126--2140},
  year={2022}
}

@inproceedings{qin2019counterfactual,
  title={Counterfactual story reasoning and generation},
  author={Qin, Lianhui and Bosselut, Antoine and Holtzman, Ari and Bhagavatula, Chandra and Clark, Elizabeth and Choi, Yejin},
  booktitle={Proceedings of the 2019 Conference on Empirical Methods in Natural Language Processing and the 9th International Joint Conference on Natural Language Processing (EMNLP-IJCNLP)},
  pages={5043--5053},
  year={2019}
}

@article{thai2025swe,
  title={SWE-EVO: Benchmarking Coding Agents in Long-Horizon Software Evolution Scenarios},
  author={Thai, Minh VT and Le, Tue and Manh, Dung Nguyen and Nhat, Huy Phan and Bui, Nghi DQ},
  journal={arXiv preprint arXiv:2512.18470},
  year={2025}
}

@article{xiong2026autoresearchbench,
  title={AutoResearchBench: Benchmarking AI Agents on Complex Scientific Literature Discovery},
  author={Xiong, Lei and Luo, Kun and Xia, Ziyi and Zhang, Wenbo and Yao, Jin-Ge and Liu, Zheng and Shao, Jingying and Chen, Jianlyu and Qian, Hongjin and Yang, Xi and others},
  journal={arXiv preprint arXiv:2604.25256},
  year={2026}
}

@article{zheng2025survey,
  title={A survey of process reward models: From outcome signals to process supervisions for large language models},
  author={Zheng, Congming and Zhu, Jiachen and Ou, Zhuoying and Chen, Yuxiang and Zhang, Kangning and Shan, Rong and Zheng, Zeyu and Yang, Mengyue and Lin, Jianghao and Yu, Yong and others},
  journal={arXiv preprint arXiv:2510.08049},
  year={2025}
}

@inproceedings{li2025generation,
  title={From generation to judgment: Opportunities and challenges of llm-as-a-judge},
  author={Li, Dawei and Jiang, Bohan and Huang, Liangjie and Beigi, Alimohammad and Zhao, Chengshuai and Tan, Zhen and Bhattacharjee, Amrita and Jiang, Yuxuan and Chen, Canyu and Wu, Tianhao and others},
  booktitle={Proceedings of the 2025 Conference on Empirical Methods in Natural Language Processing},
  pages={2757--2791},
  year={2025}
}

@inproceedings{xu2025kodcode,
  title={Kodcode: A diverse, challenging, and verifiable synthetic dataset for coding},
  author={Xu, Zhangchen and Liu, Yang and Yin, Yueqin and Zhou, Mingyuan and Poovendran, Radha},
  booktitle={Findings of the Association for Computational Linguistics: ACL 2025},
  pages={6980--7008},
  year={2025}
}

@article{xu2025toucan,
  title={Toucan: Synthesizing 1.5 m tool-agentic data from real-world mcp environments},
  author={Xu, Zhangchen and Soria, Adriana Meza and Tan, Shawn and Roy, Anurag and Agrawal, Ashish Sunil and Poovendran, Radha and Panda, Rameswar},
  journal={arXiv preprint arXiv:2510.01179},
  year={2025}
}

@inproceedings{chen2025scienceagentbench,
  title={Scienceagentbench: Toward rigorous assessment of language agents for data-driven scientific discovery},
  author={Chen, Ziru and Chen, Shijie and Ning, Yuting and Zhang, Qianheng and Wang, Boshi and Yu, Botao and Li, Yifei and Liao, Zeyi and Wei, Chen and Lu, Zitong and others},
  booktitle={International Conference on Learning Representations},
  volume={2025},
  pages={96934--96990},
  year={2025}
}

@article{chern2023factool,
  title={FacTool: Factuality Detection in Generative AI--A Tool Augmented Framework for Multi-Task and Multi-Domain Scenarios},
  author={Chern, I and Chern, Steffi and Chen, Shiqi and Yuan, Weizhe and Feng, Kehua and Zhou, Chunting and He, Junxian and Neubig, Graham and Liu, Pengfei and others},
  journal={arXiv preprint arXiv:2307.13528},
  year={2023}
}

@inproceedings{min2023factscore,
  title={Factscore: Fine-grained atomic evaluation of factual precision in long form text generation},
  author={Min, Sewon and Krishna, Kalpesh and Lyu, Xinxi and Lewis, Mike and Yih, Wen-tau and Koh, Pang and Iyyer, Mohit and Zettlemoyer, Luke and Hajishirzi, Hannaneh},
  booktitle={Proceedings of the 2023 Conference on Empirical Methods in Natural Language Processing},
  pages={12076--12100},
  year={2023}
}

@article{wei2026agentic,
  title={Agentic reasoning for large language models},
  author={Wei, Tianxin and Li, Ting-Wei and Liu, Zhining and Ning, Xuying and Yang, Ze and Zou, Jiaru and Zeng, Zhichen and Qiu, Ruizhong and Lin, Xiao and Fu, Dongqi and others},
  journal={arXiv preprint arXiv:2601.12538},
  year={2026}
}

@inproceedings{gupta2025llms,
  title={LLMs for Experiment Design in Scientific Domains: Are We There Yet?},
  author={Gupta, Rushil and Hartford, Jason and Liu, Bang},
  booktitle={ICML 2025 Generative AI and Biology (GenBio) Workshop},
  year={2025}
}

@article{ye2025limo,
  title={Limo: Less is more for reasoning},
  author={Ye, Yixin and Huang, Zhen and Xiao, Yang and Chern, Ethan and Xia, Shijie and Liu, Pengfei},
  journal={arXiv preprint arXiv:2502.03387},
  year={2025}
}

@article{chen2026towards,
  title={Towards reasoning era: A survey of long chain-of-thought for reasoning large language models},
  author={Chen, Qiguang and Qin, Libo and Liu, Jinhao and Peng, Dengyun and Guan, Jiannan and Wang, Peng and Hu, Mengkang and Zhou, Yuhang and Gao, Te and Che, Wanxiang},
  journal={Science China Information Sciences},
  volume={69},
  number={6},
  pages={161101},
  year={2026},
  publisher={Springer}
}

@article{li2025system,
  title={From system 1 to system 2: A survey of reasoning large language models},
  author={Li, Zhong-Zhi and Zhang, Duzhen and Zhang, Ming-Liang and Zhang, Jiaxin and Liu, Zengyan and Yao, Yuxuan and Xu, Haotian and Zheng, Junhao and Wang, Pei-Jie and Chen, Xiuyi and others},
  journal={arXiv preprint arXiv:2502.17419},
  year={2025}
}

@inproceedings{li2025small,
  title={Small models struggle to learn from strong reasoners},
  author={Li, Yuetai and Yue, Xiang and Xu, Zhangchen and Jiang, Fengqing and Niu, Luyao and Lin, Bill Yuchen and Ramasubramanian, Bhaskar and Poovendran, Radha},
  booktitle={Findings of the Association for Computational Linguistics: ACL 2025},
  pages={25366--25394},
  year={2025}
}

@inproceedings{zhang2026quest,
  title={The quest for efficient reasoning: A data-centric benchmark to cot distillation},
  author={Zhang, Ruichen and Khan, Rana Muhammad Shahroz and Tan, Zhen and Li, Dawei and Wang, Song and Chen, Tianlong},
  booktitle={International Conference on Learning Representations},
  volume={2026},
  pages={94555--94578},
  year={2026}
}

@inproceedings{dziri2022origin,
  title={On the origin of hallucinations in conversational models: Is it the datasets or the models?},
  author={Dziri, Nouha and Milton, Sivan and Yu, Mo and Zaiane, Osmar R and Reddy, Siva},
  booktitle={Proceedings of the 2022 Conference of the North American Chapter of the Association for Computational Linguistics: Human Language Technologies},
  pages={5271--5285},
  year={2022}
}

@article{ji2023survey,
  title={Survey of hallucination in natural language generation},
  author={Ji, Ziwei and Lee, Nayeon and Frieske, Rita and Yu, Tiezheng and Su, Dan and Xu, Yan and Ishii, Etsuko and Bang, Ye Jin and Madotto, Andrea and Fung, Pascale},
  journal={ACM computing surveys},
  volume={55},
  number={12},
  pages={1--38},
  year={2023},
  publisher={ACM New York, NY}
}

@article{kalai2025language,
  title={Why language models hallucinate},
  author={Kalai, Adam Tauman and Nachum, Ofir and Vempala, Santosh S and Zhang, Edwin},
  journal={arXiv preprint arXiv:2509.04664},
  year={2025}
}

\appendix

\section{Prompt Examples for Contextual Causality Tasks}
\label{app:prompt-examples}
Examples of prompts for representative contextual causality tasks are shown below. We use ESL \cite{caselli2017event} for ECI, CRAB \cite{romanou2023crab} for causal strength assessment, e-CARE \cite{du2022care} for causal judgment and causal explanation generation, Who\&When \cite{zhang2025agent} for causal attribution, FinCausal2020 \cite{mariko2020financial} for causal detection and span extraction, and METER \cite{li2026meter} for effect prediction and counterfactual reasoning.

\begin{tcolorbox}[
  enhanced,
  breakable,
  colback=blue!10!white,
  colframe=gray!70,
  boxrule=0.8pt,
  arc=3pt,
  left=6pt,
  right=6pt,
  top=6pt,
  bottom=6pt,
  width=\linewidth
]
\small
\setlength{\parindent}{0pt}

\textbf{Task:} Event Causal Identification

\textbf{Prompt example from ESL \cite{caselli2017event}:}

\textit{Jenkin, who was ``arrested'' at Millom Pier at around 9.35am, is also facing a charge of ``causing'' unnecessary suffering to an animal in relation to the dog's death.}

\textit{\textbf{Is there a causal relationship between ``causing'' and ``arrested''?}}

\vspace{0.6em}
\textbf{Task:} Causal Strength Assessment

\textbf{Prompt example from CRAB \cite{romanou2023crab}:}

\textit{Article A: From Mexico in the north to Chile in the south, the region's constantly see-sawing political map once again resembles that of the early 2000s, when a so-called ``pink tide'' of left-leaning governments washed over it. But analysts say this time is different: the trend is driven by pragmatism rather than ideology.}

\textit{Article B: In the 2000s, a wave of left-wing electoral success swept Latin America, with a varied and complex turn to more progressive economic or social policies. In the 2010s, a right-wing political and conservative trend countered this tide, in clear reaction to the advances of progressives, and the right has dominated politics in the region for the past decade.}

\textit{Event 1: A ``pink tide'' of left-leaning governments swept Latin America in the early 2000s. Event 2: In the 2010s, a right-wing political and conservative trend countered this tide. }

\textit{\textbf{What is the causality score between Event 1 and Event 2 from 0 to 100? Score above 80: Event 1 is definitely responsible for Event 2. Score between 50--80: Event 1 might have been responsible for Event 2. Score lower than 50: Events are somehow related but definitely not causally related.}}

\vspace{0.6em}
\textbf{Task:} Causal Judgment

\textbf{Prompt example from e-CARE \cite{du2022care}:}

\textit{Input Event: The man fell unconscious. Question: Please select the cause of the input event from the following options. Option 1: The assailant struck the man in the head. Option 2: The assailant took the man's wallet.}

\textit{\textbf{Which option is the cause of the input event?}}

\vspace{0.6em}
\textbf{Task:} Causal Attribution

\textbf{Prompt example from Who\&When \cite{zhang2025agent}:}

\textit{Context: The following is a failed multi-agent task-solving process. The system consists of three agents: Excel\_Expert, BusinessLogic\_Expert, and DataVerification\_Expert.}

\textit{Original task: This spreadsheet contains a list of clients for a retractable awning company. Each client has ordered a new awning for the back of their house within the last 90 days. The company makes different designs depending on whether the awning is made to block sunrises or sunsets. In this region, houses with odd-numbered street addresses face east, and houses with even-numbered street addresses face west. How many of these clients will be receiving the sunset awning design?}

\textit{Agent system prompts:}

\textit{Excel\_Expert: Your role Excel\_Expert: Expert in analyzing and processing data from Excel files. Task and skill instructions -- Task Description: You will be responsible for scrutinizing, interpreting, and managing data from Excel files. The role requires a deep understanding of various Excel functions, including but not limited to pivot tables, data visualization, data cleaning, and advanced formulas. Skill Description: You possess exceptional skills in Excel, enabling you to manipulate, analyze, and derive meaningful insights from complex datasets. You have the ability to automate repetitive tasks using VBA (Visual Basic for Applications) and generate comprehensive reports. Your expertise includes proficiency in handling large datasets, performing statistical analysis, and utilizing Excel's extensive array of tools to streamline data processing.}

\textit{DataVerification\_Expert: Your role DataVerification\_Expert specializes in employing Bing Search API for powerful online search capabilities. With an adept skill set in navigating, parsing, and analyzing search results, this expert seamlessly extracts precise information from an array of text content. A critical part of their role involves vigilantly verifying the accuracy of this information, ensuring the use of reliable, trustworthy data. Task and skill instructions -- You will utilize the Bing Search API to execute specialized search queries, combing through the vast data available on the internet. Your skill in parsing and analyzing search results will be crucial in distilling the vast array of information to the most relevant details. When it comes to extracting specific information from text content, your ability to pinpoint and retrieve exact data points is paramount. An essential component of your work is verifying the accuracy of information, which requires a keen eye for detail and a robust methodology for cross-checking facts. With your combination of technical prowess using the Bing Search API and a meticulous approach to data verification, your role as the DataVerification\_Expert is essential for any operation that demands high accuracy and reliability in information gathering. Useful instructions for task-solving -- Follow the instructions provided by the user. Solve the task step by step if you need to. If a plan is not provided, explain your plan first. If the error cannot be fixed or the task is not solved even after the code executes successfully, analyze the problem, revisit your assumptions, collect any additional information you need, and consider a different approach. When you find an answer, verify the answer carefully. Include verifiable evidence in your response if possible. How to use code? Suggest Python code in a Python coding block or shell script in a sh coding block for the Computer\_terminal to execute. When using code, you must indicate the script type in the code block. Do not suggest incomplete code that requires users to modify it. The last results will not be cached, so you need to provide all necessary information in a single code block. Do not use a code block if it is not intended to be executed by the Computer\_terminal. The Computer\_terminal cannot provide any other feedback or perform any other action beyond executing the code you suggest. The Computer\_terminal cannot modify your code. Use the print function for the output when relevant. Check the execution result returned by the user. Do not ask users to copy and paste the result. If the result indicates an error, fix it and output the code again. If you want the Computer\_terminal to save the code in a file before executing it, put \# filename: <filename> inside the code block as the first line.}

\textit{BusinessLogic\_Expert: Your role BusinessLogic\_Expert is an expert in understanding and applying business rules to data. This role involves understanding complex business requirements and translating them into actionable data logic aligned with organizational goals. Task and skill instructions -- Your task involves analyzing business requirements and interpreting them into specific business rules that can be implemented within a data framework. You must possess a deep understanding of business processes and their relationship to data management and manipulation. The ability to apply these rules accurately to data sets is crucial for ensuring the integrity and efficiency of business operations. Key skills include critical thinking, attention to detail, problem-solving, and proficiency with various data tools and platforms. You should be adept at both written and verbal communication to effectively convey business logic to stakeholders and team members. Additional knowledge in SQL, database design, and data analysis techniques will further enhance your capability in this role. By leveraging your expertise, you will contribute significantly to optimizing business operations through precise and effective data logic application.}

\textit{Execution trace:}

\textit{Step 0 -- Excel\_Expert: The Excel\_Expert receives the general task and the manager's plan. It loads the Excel file from \path{../2023/validation/4d51c4bf-4b0e-4f3d-897b-3f6687a7d9f2.xlsx}, identifies the \texttt{Street Address} column, extracts the first numeric substring from each address using \texttt{str.extract} with a digit-matching regular expression, converts it to an integer, filters even-numbered street addresses, and prints the number of such clients.}

\textit{Step 1 -- Computer\_terminal: The code executes successfully. The terminal outputs the spreadsheet columns: \texttt{ID}, \texttt{First Name}, \texttt{Last Name}, \texttt{Street Address}, \texttt{City}, \texttt{Zip}, \texttt{Phone}, \texttt{Email}, and \texttt{Carrier Pigeon}. The terminal then outputs \texttt{4}.}

\textit{Step 2 -- BusinessLogic\_Expert: The BusinessLogic\_Expert verifies that the column name is \texttt{Street Address}. It repeats the same reasoning: load the file, extract street numbers from \texttt{Street Address}, identify even-numbered addresses, and count the clients with even-numbered addresses. It provides essentially the same Python code and asks the terminal to run it again.}

\textit{Step 3 -- Computer\_terminal: The code executes successfully again and outputs \texttt{4}.}

\textit{Step 4 -- DataVerification\_Expert: The DataVerification\_Expert accepts the executed result. It states that the street-address column was correctly identified, the street numbers were accurately extracted, clients with even-numbered addresses were identified, and the final count of clients receiving the sunset awning design is \texttt{4}.}

\textit{Step 5 -- DataVerification\_Expert: TERMINATE.}

\textit{\textbf{The above multi-agent system produced an incorrect final answer. Which agent caused the task failure, and at which step did the decisive error occur?}}

\vspace{0.6em}
\textbf{Task:} Causal Explanation Generation

\textbf{Prompt example from e-CARE \cite{romanou2023crab}:}

\textit{Cause: The assailant struck the man in the head. Effect: The man fell unconscious.}

\textit{\textbf{Why can the cause lead to the effect?}}

\vspace{0.6em}
\textbf{Task:} Causal Detection and Span Extraction

\textbf{Prompt example from FinCausal2020 \cite{mariko2020financial}:}

\textit{
Text: Things got worse when the Wall came down. GDP fell 20\% between 1988 and 1993. There were suddenly hundreds of thousands of unemployed in a country that, under Communism, had had full employment.}

\textit{\textbf{Does the text express a causal relation? If yes, extract the cause span and the effect span.}
}

\vspace{0.6em}
\textbf{Task:} Effect Prediction

\textbf{Prompt example from METER \cite{li2026meter}:}

\textit{On January 12, 2025, patient Robert Chen (Male, 58) was admitted to the Cardiac Care Unit at City General Hospital. Coronary angiography confirmed a thrombotic occlusion in the proximal Left Anterior Descending (LAD) artery, causing the acute myocardial ischemia observed on the ECG. Regarding treatment options, clinical protocols define Percutaneous Coronary Intervention (PCI) as the standard procedure for mechanically recanalizing occluded vessels to re-establish blood flow. Furthermore, pathology reports indicate significant myocardial necrosis, which is time-dependent and developed progressively over the four-hour duration following his initial presentation at 09:00, during which the occlusion remained untreated.
}

\textit{\textbf{Question: If Percutaneous Coronary Intervention is performed, what is the expected outcome for the patient?}}

\vspace{0.6em}
\textbf{Task:} Counterfactual Reasoning

\textbf{Prompt example from METER \cite{li2026meter}:}

\textit{On January 12, 2025, patient Robert Chen (Male, 58) was admitted to the Cardiac Care Unit at City General Hospital. Coronary angiography confirmed a thrombotic occlusion in the proximal Left Anterior Descending (LAD) artery, causing the acute myocardial ischemia observed on the ECG. Regarding treatment options, clinical protocols define Percutaneous Coronary Intervention (PCI) as the standard procedure for mechanically recanalizing occluded vessels to re-establish blood flow. Furthermore, pathology reports indicate significant myocardial necrosis, which is time-dependent and developed progressively over the four-hour duration following his initial presentation at 09:00, during which the occlusion remained untreated.
}

\textit{\textbf{Would the extent of myocardial necrosis have been reduced if the occlusion had been detected during the initial screening?}
}
\end{tcolorbox}

\section{Literature Collection Scope}
\label{app:literature-collection}

We collected literature according to the role of each study in this survey. For comparison surveys, we considered works from major NLP venues such as EMNLP and NAACL, broader AI and data mining venues such as KDD, AI-related journals such as IEEE Transactions on Artificial Intelligence and ACM Computing Surveys, as well as highly relevant preprints that have not yet been formally published. For method-oriented studies, we primarily considered papers from major NLP venues, including ACL, EMNLP, NAACL, EACL, and COLING, while also including highly relevant studies from broader AI venues. For datasets and benchmarks, we mainly included representative and widely used resources relevant to the contextual causality tasks discussed in this survey.

\end{document}